\documentclass[runningheads]{llncs}
\usepackage{graphicx}
\usepackage{times}
\usepackage{epsfig}
\usepackage{breqn}
\usepackage{float}
\usepackage{amsmath}
\usepackage{amssymb}
\begin{document}
%
\title{Low Cost Embedded Vision System For Location And Tracking Of A Color Object}
%
%
\author{Diego Ayala \inst{1,2} \and
Danilo Chavez\inst{2}\and
Leopoldo Altamirano Robles\inst{1}}
\authorrunning{D. Ayala et al.}
%
\institute{Instituto Nacional de Astrofísica, Óptica y Electrónica (INAOE), Puebla, Mexico
    \email{ayala.diego@inaoe.edu.mx},
	\email{robles@inaoep.mx}
\and
Escuela Politécnica Nacional, Quito, Ecuador\\
 \email{danilo.chavez@epn.edu.ec}
}

\maketitle              
\begin{abstract}
This paper describes the development of an embedded vision system for detection, location, and tracking of a color object; it makes use of a single 32-bit microprocessor to acquire image data, process, and perform actions according to the interpreted data.  The system is intended for applications that need to make use of artificial vision for detection, location and tracking of a color object and its objective is to have achieve at  reduced terms of size, power consumption, and cost.

\end{abstract}
	\section{Introduction}

Several fields such as industrial, military, scientific and civil have chosen to make use of computer vision in order to recognize the existence of objects and their location, among other features; most of these systems need a personal computer and the execution of the software that processes the image data.
Applications such as unmanned vehicle systems, autonomous robots, among others, have limitations of space, consumption, robustness and weight, making the use of a personal computer to be impractical, or requiring complex and expensive methods to transmit the image to a fixed station that processes the image and re-transmit the data interpretation.
Reduced size systems have been implemented on commercial boards such as the so called Cognachrome Vision System but it requires an external camera connected to a RCA protocol adapter \cite{Sargent}, yet another similar work was made by a team at Carnegie Mellon University \cite{Rowe} but it lacks of an embedded user interface and costs more than the development proposed in this document, which is a compact embedded vision system, lightweight, with a low power consumption, and written in widely used C/C++ language it handles: the image acquisition, processing, and a user interface altogether on a board and camera that are roughly USD 90 in price. 
The proposed system intends to be a cheaper, easy to replicate, and yet a viable and modern alternative to the ones researched by A. Rowe et al. \cite{Rowe}, R. Sargent et al. \cite{Sargent}; its further development could contribute to applications that require to detect, locate and/or track a color object and have strong limitations in: size, power consumption and cost.
\section{Capturing and display of the image}
This project uses a HY-Smart STM32 development board, it includes a STM32F103 microcontroller to process data, it gets the image from an OV7725 camera that is configured in RGB565 format, with a QVGA(320x240) resolution. It also includes a touch screen in which the target object can be selected, its color defines the threshold that is used to create a binary image in the process of artificial vision known as segmentation. After the segmentation is done, an algorithm recognizes the contour of the image and its center, once located, a PID algorithm commands 2 servos (pan, tilt) in order to track the objective.	
\subsection{Image acquisition}

\begin{figure}[h]
	\centering
	\includegraphics[width=0.95\linewidth]{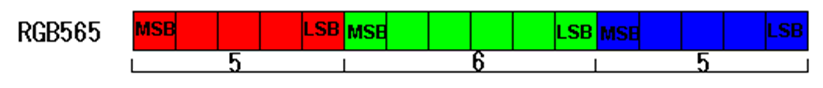}
	\caption{RGB565 format contains 16 bits of information per pixel.}
	\label{fig:rgb565}
\end{figure}

The project make use of the OV7725 camera in a RGB565 format, which employs 2 bytes per pixel, allocating 5 bits for red, 6 for green, and 5 for blue as seen on Fig.~\ref{fig:rgb565}. An individual frame has 320x240 pixels of information which are constantly been sent to a FIFO memory named AL422B; the microcontroller accesses this data when required, instead of receiving periodic interruptions from the camera.
\subsection{Displaying the image}
\begin{figure}[h]
	\centering
	\includegraphics[width=0.95\linewidth]{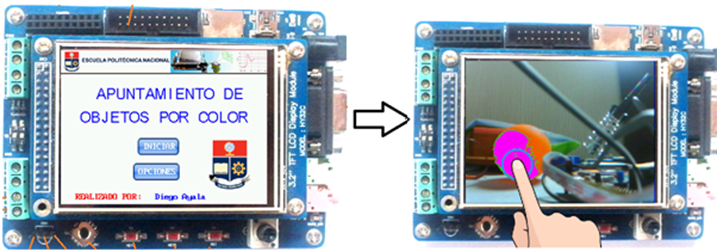}
	\caption{Image of the interface and user interaction with the screen.}
	\label{fig:lcd}
\end{figure}

A TFT-LCD screen of 320x240 pixels displays the image, it is operated by the SSD1289 integrated circuit that communicates with the microcontroller through 8080 parallel protocol. A resistive film above the screen, in conjunction to the XPT2046 integrated circuit, locates the position of a single pressure point on the screen and sends the data via SPI interface.
The microcontroller has a peripheral block called FSMC (Flexible Static Memory Controller) which allows it to communicate with external memories meeting the timing requirements, previously some parameters must be set: the type of memory to be read (SRAM, ROM, NOR Flash, PSRAM), data bus width (8 or 16 bits), the memory bank to be used, waiting times, among other features.
The use of the abovementioned integrated circuits allows the microcontroller to seamlessly read and write the camera and display respectively, and allows a user interface as depicted on Fig.~\ref{fig:lcd}, although the display could be discarded in order to increase frames per second, and decrease cost, and weight.
\section{Image processing}
The 76800 pixels contained in each frame need to be processed in order to detect and locate the color object, this task is described as segmentation of the image. Once located a PID controller centers the field of view of the camera on the center of the region of interest.
\subsection{Image segmentation}
\begin{figure}[h]
	\centering
	\includegraphics[width=0.65\linewidth]{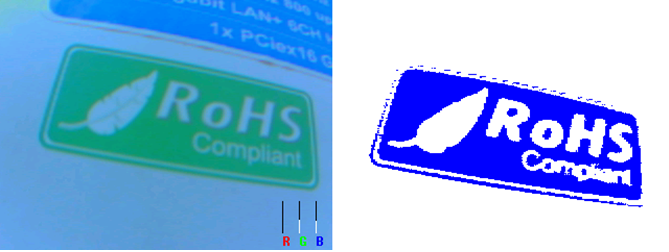}
	\caption{On the left the original image, on the right the binary image.}
	\label{fig:segmentated}
\end{figure}

The segmentation consists on separating the region of interest in the image based on the chosen color. As each pixel is obtained from the camera, it is compared with a threshold value for each channel, and the result is stored into a binary image.
The binary image is allocated in memory as an array of 2400 X 32bit numbers where each bit is a pixel of the binary image (see Fig.~\ref{fig:segmentated}), the color boundaries can be selected via the interface, two approaches are being considered: RGB color space and normalized RGB.
\subsubsection{RGB color space}
Maximum, and minimum boundaries are set for each of the three color components (red, green and blue), if the scanned pixel is within the 3 ranges, then it is stored as “one” in the binary image, otherwise is a “zero”. This computing is fairly fast, achieving 10,2fps; but a disadvantage arise in the event of a change of illumination, for instance if light intensity is decreased the red, blue and green components vary in proportion to this change, and can get out of the threshold, the same occurs with an increase in light intensity. 
\subsubsection{Normalized RGB}
\begin{figure}[h]
	\centering
	\includegraphics[width=0.65\linewidth]{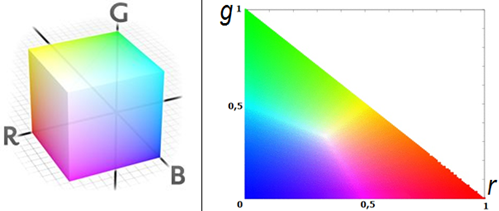}
	\caption{On the left RGB color space, on the rg chromaticity space.}
	\label{fig:spaces}
\end{figure}

\begin{figure}[h]
	\centering
	\includegraphics[width=0.65\linewidth]{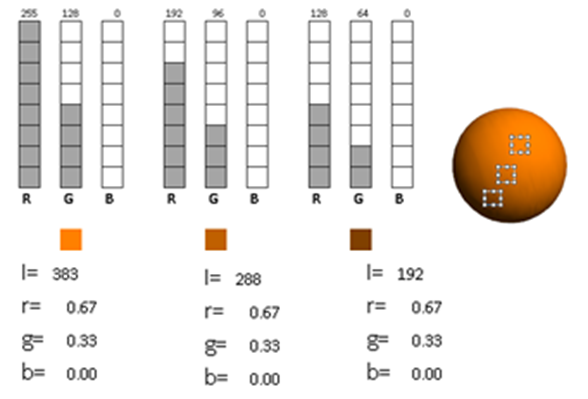}
	\caption{Three different color pixels are chosen from an orange sphere, rgb components are computed on each case.}
	\label{fig:spacestest}
\end{figure}

In this color space, instead of using directly each RGB component, the proportion \textit{rgb} is calculated by dividing the luminance \textit{I} of every single pixel \cite{Balkenius}. 
\begin{align}
	I &=R+G+B 	\\
	r  &=R/I,g=G/I,b=B/I \
\end{align}
As the name suggests, in the normalized RGB the summation of rgb components equals to one, due to this only r and g values are calculated to attain the hue information, this results in the rg chromaticity space seen on Fig.~\ref{fig:spaces}, which is bi-dimensional and theoretically invariant to changes of illumination.
The invariance mentioned can be noted in the example given on Fig.~\ref{fig:spacestest} where three pixels o an orange sphere are evaluated, the R component varies along this points in nearly half of its value, selecting a threshold in RGB color space would neglect a considerable part of the sphere due to the large of R. However, while the RGB values vary, their proportions with respect of the intensity (I) keep the same, thus the values of rgb are invariant to the distinct illumination levels. An experimental result is documented on Figure~\ref{fig:testseg}
\subsection{Description of the region of interest}
To analyze the data incoming from the image segmentation, an algorithm demarks the contour of the group of contiguous pixels in the binary image, once this is done it establish the upper, lower, rightmost, and leftmost limits, and also both horizontal and vertical location of the center of the object.
The algorithm starts by scanning the binary image from the top left corner, to the right and downwards until it find a line of contiguous pixels, if it exceeds a preset width then it finds the rightmost pixel of the grouping and proceeds to find the pixels of the contour.
\begin{figure}[h]
	\centering
	\includegraphics[width=0.65\linewidth]{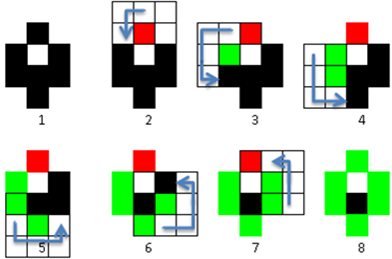}
	\caption{Example of contour recognition for a group of pixels.}
	\label{fig:roi}
\end{figure}Fig.~\ref{fig:roi} shows how the algorithm runs along the contour of a sector of detected pixels , the process find the initial line ( in this case the first line is a single pixel width, and is shown in red color), from here the contour path begins, as a rule the algorithm begins to search for the next valid counter-clockwise pixel in a 3x3 matrix, initiating the inspection from the next position to the last sensed pixel. To continue the contour detection , the center of the next matrix is at the position of pixel detected earlier. Whenever a contour pixel is detected it is evaluated to update the upper, lower, rightmost, and leftmost limits of the grouping of pixels. The contour inspection stops once the initial pixel is reached.
\subsection{Tracking}
\begin{figure}[h]
	\centering
	\includegraphics[width=0.4\linewidth]{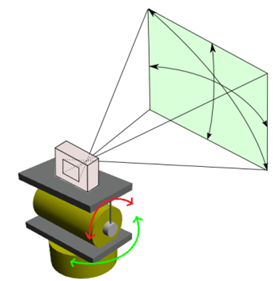}
	\caption{Representation of the Pan-Tilt platform holding the camera.}
	\label{fig:servos}
\end{figure}

The camera is located in the top of a Pan-Tilt platform represented on Fig.~\ref{fig:servos}, in order to perform the tracking movements two servo motors are installed: HS-785HB servo motor (located at the bottom of the platform) and the HS-645MG (located on top).
The control algorithm that governs the movements of both servos is a proportional-integral (PI) controller. Although the system is composed of servomotors, a modeling of the dynamic response was made in order to represent the platform with the camera installed, the result is a first-order transfer function (Gp) whose parameters: gain (K) and time constant (\(\tau\)) are experimentally acquired.

\subsubsection{Closed-loop analysis }
\begin{figure}[h]
	\centering
	\includegraphics[width=0.75\linewidth]{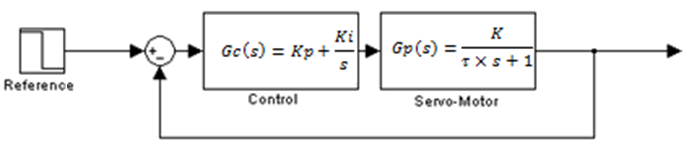}
	\caption{Simplified block diagram of the closed loop system.}
	\label{fig:loop}
\end{figure}

The PI controller was chosen to eliminate the offset error, a block representation of the system in continuous time domain is depicted on Fig.~\ref{fig:loop}.

The vision system gets feedback through the position of the detected object relative to the camera, the controller will govern the PWM that moves the servomotors in order to move the camera so its center is aligned with the object’s center, both variables are relative to the angular location of the camera, so the error signal is invariant to the absolute angular location of the platform, in other words the error is just measured as the difference between the detected position and the center of the camera’s field of view.
The block diagram of  Figure ~\ref{fig:loop} can be simplified into a single block to obtain the following transfer function representing the whole system:
\begin{align}
	Tf(s)& =\frac{Gp(s)\times Gc(s)}{Gp(s)\times Gc(s)+1}	\
\end{align}
Further calculation leads to the following second order transfer function:
\begin{align}
	Tf(s)=\frac{s\times (Kp\times K)+Ki\times K}{s^2\times \tau+s\times (1+Kp\times K)+Ki\times K}	\
\end{align}
The poles of the closed loop system are given by the roots of the polynomial on the denominator\cite{Ogata}:
\begin{align}
	p(s)=s^2+s\times (2 \xi  \omega_n)+\omega_n^2	\\
	ts=\frac{4}{\xi\times\omega_n}\
\end{align}
The damping factor (\(\xi\)) and natural frequency (\(\omega_n\)) determine the percentage overshoot (PO) and settling time (ts) present in the transient response of a step input [4]:
\begin{align}
	PO=100\% \times e^{\frac{-\xi\times \pi}{\sqrt{1-\xi^2}}}	\
\end{align}
The controller's constants (Kp and Ki) can be solved performing the replacements required in Equations (4), (5), (6) and (7).
\begin{align}
	Kp=\frac{1}{K}\times(\frac{8\times \tau}{t_s-1})\
\end{align}
\begin{align}
Ki=\frac{1}{K}\times \frac{16 \times \tau}{(t_s)^2\times \frac{ln(\frac{PO}{100\%})^2}{\pi^2+ln(\frac{PO}{100\%})^2}}\
\end{align}
\subsubsection{Discretization of the controller }
The described PI controller (Gc) is expressed in continuous time-domain. To make an algorithm executable by the microcontroller the PI controller must be discretized. 
This is achieved by the bilinear transformation function Equation (10) which transforms the transfer function from continuous time domain to the discrete time domain \cite{mit}:
\begin{align}
	G(z)= G(s)|_{s=\frac{2(z-1)}{T(z+1)}}\
\end{align}
Where T is the sampling time. Applying the transformation to the PI controller we obtain the following transfer function:
\begin{align}
	Gc(z)=Kp+\frac{Ki\times T(z+1)}{2(z-1)} =\frac{U(z)}{E(z)}\
\end{align}
The discrete time domain controller can be expressed in a single line mathematical operation, thus finally obtaining the control law:
\begin{multline}
	U_{[k]} =U_{[k-1]} +E_{[k-1]} \times (Ki\times \frac{T}{2}-Kp)+E_{[k]} \times (Ki\times \frac{T}{2}+Kp) \
\end{multline}
Where U[k] is the instantaneous value of the control action (value servomotor PWM pulse), U[k-1] the previous value, the error  E[k] is the difference between the center of the camera and the center of the object; the controller updates this values every time a frame is acquired which occurs at sampling a time T, on the other hand Kp and Ki are constants determined by the equations (8) and (9) respectively.
\section{Tests and results}
\subsection{Image segmentation }
\begin{figure}[h]
	\centering
	\includegraphics[width=0.5\linewidth]{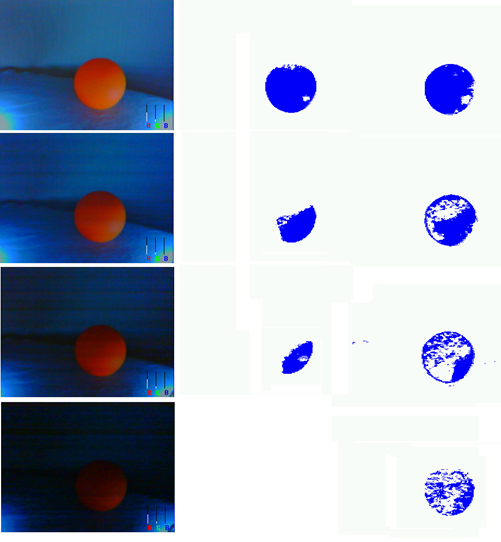}
	\caption{(Left) An orange sphere is illuminated at 4 poor intensity scenarios. (Center) Results of RGB565 segmentation. (Right) Results of the rg chromaticity segmentation. (the color to detect was chosen during the highest level of illumination for both types of segmentation)}
	\label{fig:testseg}
\end{figure}

The transformation to the rg chromaticity space gives better segmentation results as can be seen on Fig.~\ref{fig:testseg}, nevertheless the calculation of each pixel takes more time than RGB color space, resulting in a relatively slower frame rate of 10fps. For the rest of this tests rg chromaticity segmentation is chosen, as it is more reliable.
\subsection{Detection of distinct color objects}
\begin{figure}[h]
	\centering
	\includegraphics[width=0.5\linewidth]{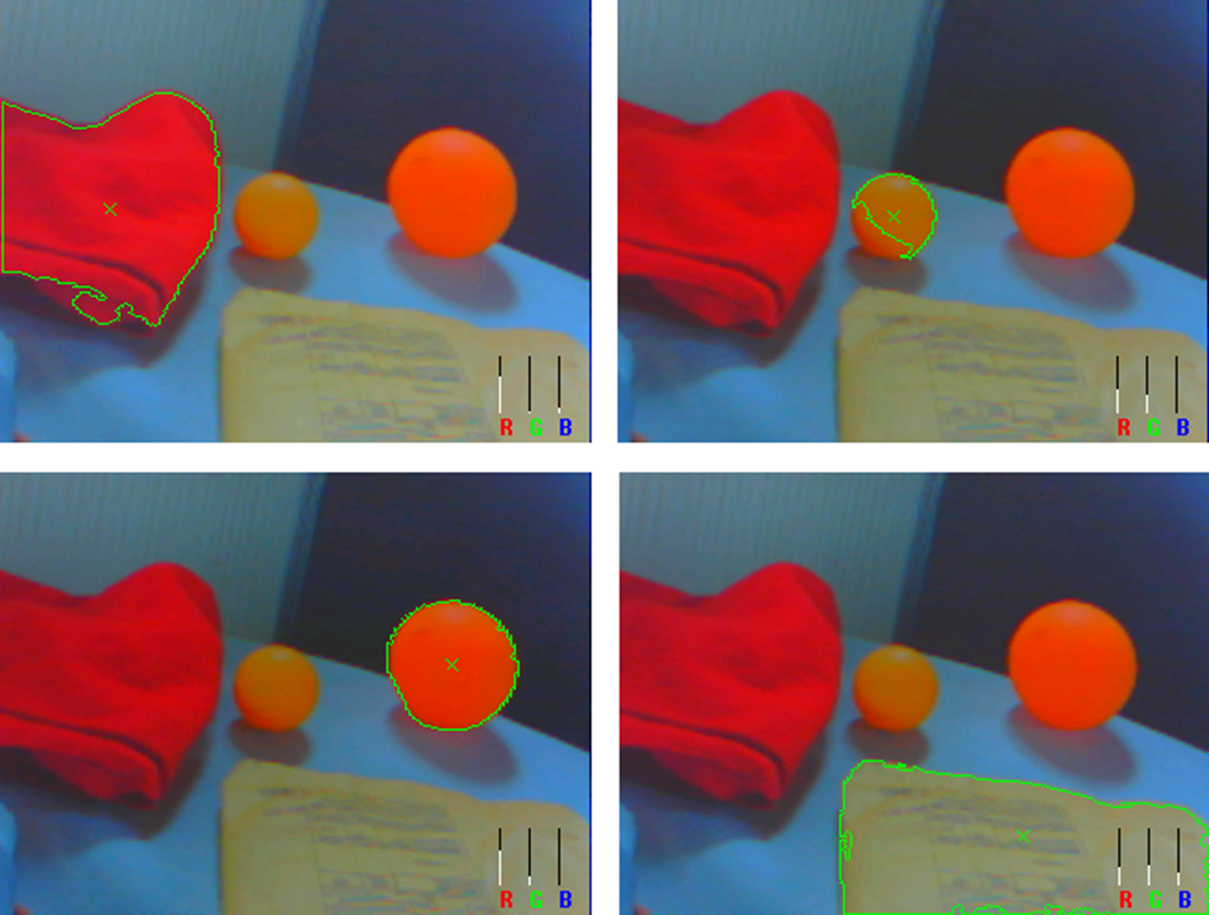}
	\caption{Different objects of similar color are being recognized.}
	\label{fig:testobjects}
\end{figure}

Figure~\ref{fig:testobjects} shows distinct objects whose color is not much different from each other. The image has: a red cloth, a small yellow sphere, a large orange sphere, and an envelope of pale yellow. In each of the 4 experiments the respective color is selected, and the location of the objects is performed appropriately.

\subsection{Location and tracking}
\begin{figure}[h]
	\begin{center}
		\includegraphics[width=0.8\linewidth]{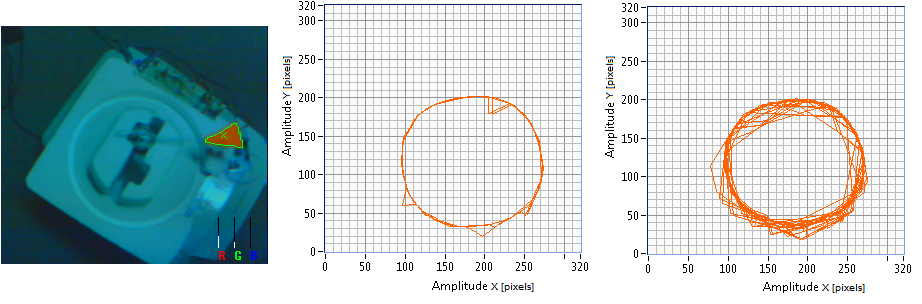}
		\caption{(Left) An Orange triangle making a clockwise motion. (Center) Location at 3820ms per revolution, (Right). Location at 1108ms per revolution.}
		\label{fig:testlocation}
	\end{center}
\end{figure}

An orange object was attached to a coupled shaft which a circular motion, similar to a clock (Figure~\ref{fig:testlocation}). Tracking is disabled and the object's position is measured as pixels, which returns a circle with a mean radius of \textit{R=87.57} pixels and a standard deviation \textit{$\sigma$ =8.69}, this along with the observed plot suggests that location data incorporates some glitches that can be addressed to partial unrecognized regions thus computing a different centroid. 

\begin{figure}[h]
	\centering
	\includegraphics[width=0.6\linewidth]{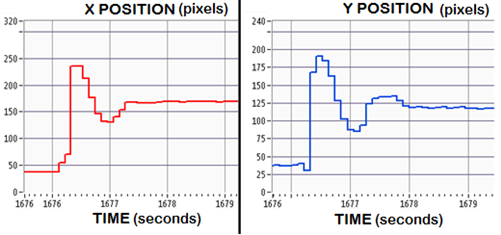}
	\caption{Tracking an object from a corner to the center of the field of view.}
	\label{fig:testtrack}
\end{figure}

\begin{figure}[h]
	\centering
	\includegraphics[width=0.6\linewidth]{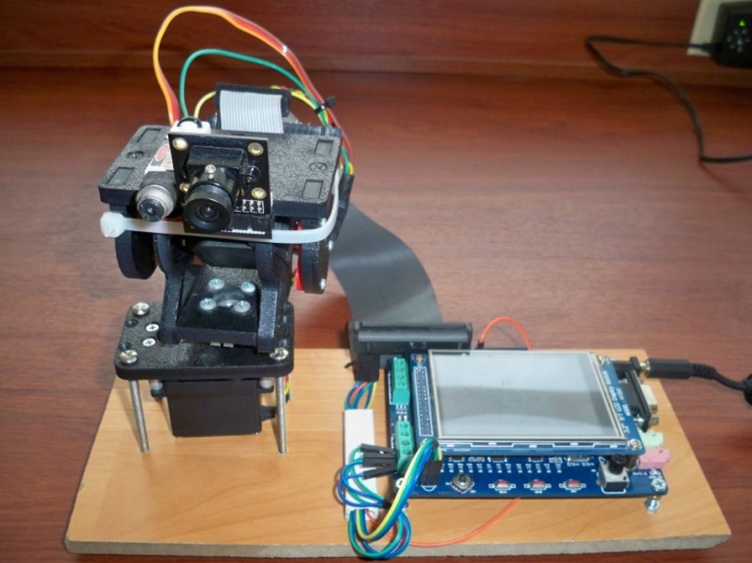}
	\caption{Physical implementation the system.}
	\label{fig:real}
\end{figure}

However this doesn't cause an instability when the servos are activated for tracking, as demonstrated in a test where a color object was chosen, brought to a corner of the camera's range of vision and then tacking was activated enabling the servomotors to move the center of the camera to the object's location, the behavior can be observed on Fig.~\ref{fig:testtrack}.
Position of the object settles to the center after roughly 1,6 seconds after tracking activation, similar results were obtained on latter tests.
The whole system is displayed on Figure~\ref{fig:real}.
\subsection{Cost and power consumption}
The data presented on Table. \ref{table:price} and \ref{table:parameter}, include the system operating with both the LCD screen, and the pan and tilt platform; even though it is totally functional without them.
	\begin{table}[H]\centering
		\begin{tabular}{|l|l|ll}
			\cline{1-2}
			\textbf{Product}                            & \textbf{Price {[}USD{]}} &  &  \\ \cline{1-2}
			HY-Smart STM32 (STM32F103VCT+TFT LCD+Board) & 59.00                    &  &  \\ \cline{1-2}
			OV7725 camera + AL422B FIFO Module                   & 30.00                    &  &  \\ \cline{1-2}
			SPT200 Direct Drive Pan \& Tilt System      & 45.99                    &  &  \\ \cline{1-2}
			\textbf{Total}                              & \textbf{134.99}          &  &  \\ \cline{1-2}
		\end{tabular}
	\caption{Price of components for the proposed Low-Cost embedded vision system}
	\label{table:price}
	\end{table}
With the listed price of 134.99 USD the system is cheaper than the one proposed in the work of A. Rowe (199 USD) \cite{Rowe}, unfortunately a comparison with the Cognachrome Vision System \cite{Sargent} is not possible as the price-tag of their system is not publicly available, on the latter a clear strength arises from the fact that the proposed system is made from already available and cheap components in the consumer market. Also it is worth noting that the system can be greatly reduced on its components being STM32F103VCT the main component at a price of 9.07 USD, which is capable enough to process image data when compared to newer architectures and solutions such as the well-known Raspberry PI, which costs 35 USD.

\begin{table}[H]\centering
	\begin{tabular}{|l|l|l|l|}
		\hline
		\textbf{Item}     & \textbf{Average} & \textbf{Max} & \textbf{Unit} \\ \hline
		Supply voltage    & 5                & 5            & V             \\ \hline
		Operating current & 200              & 1100         & mA            \\ \hline
		Frequency         & 72               & -            & MHz           \\ \hline
		Start-up time     & 500              & -            & mS            \\ \hline
		RS-232 bit-rate   & 9600           & 921600            & bps           \\ \hline
		Refresh rate   & 10.9           & -            & fps            \\\hline		
	\end{tabular}
	\caption{Electrical characteristics of the proposed Low-Cost embedded vision system}
\label{table:parameter}
\end{table}
The power consumption is 1 Watt on average, therefore it can be operated in the scope most autonomous systems; with a refreshing rate of 10.9 fps and standard RS-232 output of data, it could be easily implemented on industry processes such as: fruit classification, object location, and others. Much of the weight, power usage, and overall dimensions can be further lowered without the servomotors and LCD depending on the application.

\section{Conclusions}
The deterring effects of uncontrolled illumination are greatly diminished by the use of the rg chromaticity space enabling this  system to detect, locate, and track a colored object satisfactorily, while being low-cost (under 200USD), compact (30x13x19cm including the platform), and energy-saving (200mA on average at 5V).\\

The ability of this system to recognize chromaticity along with location data can be greatly improved in a controlled environment making it a suitable and economic option for industrial applications; according to the requirements additional work is needed for the system to locate multiple objects at the same time, and more tasks that other systems achieve running operating systems, nonetheless, this system present a significant reduction in cost, size and power consumption, which makes it viable to be fitted on small unmanned vehicles.\\

Further development of this work can be done to increase the frames per second rate, both the camera and microcontroller have newer versions in the market by the date, all the algorithms used in this work are written in C++ so they can be implemented in other systems as well and become adaptable to variety of requirements.

%
%
%

\end{document}